%% file: main.tex
\documentclass{article} %
\PassOptionsToPackage{table}{xcolor}
\usepackage{iclr2025_conference}

\input{math_commands.tex}

\usepackage{tabularx}
\usepackage{hyperref}
\usepackage{url}
\usepackage{xspace}
\usepackage{amssymb}
\usepackage{xcolor}          %
\usepackage{multirow}
\usepackage{multicol}
\usepackage{booktabs}
\usepackage{amsmath,amsthm} 
\IfFileExists{accessibility.sty}{%
  \usepackage{accessibility}%
}{}%

\newcommand{\best}[1]{\cellcolor{dreamblue!15}#1}     
\newcommand{\second}[1]{\textcolor{dreamblue}{#1}}   
\newcommand{\bestlabel}[1]{\colorbox{dreamblue!15}{\strut #1}}

\hypersetup{
  colorlinks=true,
  linkcolor=dreamblue,
  citecolor=dreamblue,
  urlcolor=dreamblue,
  filecolor=dreamblue,
}

\renewcommand{\dreamxrunhead}{DreamX}                 %
\renewcommand{\dreamxvenue}{DreamX}  %
\renewcommand{\dreamxdate}{September 2026}                  %

\title{IntBMoE: Integrating Block-Level Conditioning into Expert Composition for Full-Participation Mixture-of-Experts}

\author{%
Ran Cheng\thanks{Equal contribution.}, Longfei Xu\footnotemark[1]  \thanks{Project lead.}, Zheng Liu\footnotemark[1], Kaikui Liu, Xiangxiang Chu
\AND
{\normalfont Amap, Alibaba Group}%
}

\iclrfinalcopy %

\begin{document}

\maketitle

\input{tex/abstract}

\input{tex/other}

\bibliography{main}
\bibliographystyle{iclr2025_conference}

\appendix

\input{tex/appendix}

\end{document}

%% file: math_commands.tex
\usepackage{amsmath,amsfonts,bm}

\def\eqref#1{equation~\ref{#1}}

\def\1{\bm{1}}

\DeclareMathAlphabet{\mathsfit}{\encodingdefault}{\sfdefault}{m}{sl}
\SetMathAlphabet{\mathsfit}{bold}{\encodingdefault}{\sfdefault}{bx}{n}

\newcommand{\softmax}{\mathrm{softmax}}

\DeclareMathOperator{\TopK}{TopK}

%% file: tex/abstract.tex
\begin{abstract}

Mixture-of-Experts (MoE) scales capacity, but existing designs cannot set three quantities independently. For a single token, \textbf{participation} is how many experts contribute knowledge to its output, \textbf{execution} is how many are actually computed (compute cost), and \textbf{materialization} is how many expert-sized parameter sets must be built and stored (memory cost). Sparse routing keeps execution and materialization low, but shrinks participation: for each token, only a few experts contribute. Dense output-mixing restores full participation, but its execution grows with the number of experts. Parameter-merging keeps execution at one expert, but its materialization grows with the number of routing decisions.

We propose IntBMoE, a block-conditioned MoE that decouples all three by pairing dense expert composition with sparse block execution. Its blocks come from a small learned codebook, one per entry. At each internal layer, a lightweight hypernetwork merges all expert bases in that layer's pool into one composed expert. \textbf{Participation is full}, because every composed expert draws on the entire pool. \textbf{Execution stays sparse}, because a router sends each token to only a few blocks. \textbf{Materialization is bounded}, because the codebook, not the input, fixes how many blocks exist. Dual-Path Residual Gating (DPRG) further couples two independently composed paths through multiplicative gating.

Experiments on image classification show consistent gains over representative
sparse and dense MoE baselines. Additional experiments on
language modeling and sequential recommendation validate its generalization
beyond vision.
IntBMoE is fully deployed in AMap's generative recommendation system,
serving hundreds of millions of users under a \(60\,\mathrm{ms}\) latency
budget, with a \(2.4\%\) relative UVCTR gain in online A/B testing.
Our code is available at
\url{https://github.com/AMAP-ML/DreamX-Rec/}.

\end{abstract}

%% file: tex/other.tex
\section{Introduction}

Mixture-of-Experts (MoE) architectures have been widely adopted in language
models~\cite{lepikhin2020gshard,fedus2022switch,jiang2024mixtral,dai2024deepseekmoe},
vision models~\cite{riquelme2021scaling,fan2022m3vit}, multimodal
models~\cite{mustafa2022multimodal,xue2023raphael,li2024cumo}, and recommendation
systems~\cite{deng2025onerec,zhu2025rankmixer}. A standard MoE layer contains a
pool of experts and a router that determines how the experts process each
token. Existing designs follow three main strategies. \emph{Sparse-routing
methods} execute only a small number of selected experts~\cite{shazeer2017outrageously,lepikhin2020gshard,fedus2022switch,dai2024deepseekmoe}.
\emph{Dense output-mixing methods} execute every expert and combine their
outputs~\cite{ma2018modeling,tang2020progressive}. \emph{Parameter-merging
methods} combine the parameters of all experts into a single composite
expert and then execute it~\cite{muqeeth2023soft,zhong2024lory}.

We analyze these approaches along three dimensions. \textbf{Participation}: how many experts contribute knowledge to a
token's output? \textbf{Execution}: how many of them must actually be computed,
and so how much compute? \textbf{Materialization}: how many expert-sized
parameter sets must be built and stored (one for every distinct routing
decision), and so how much memory? In principle the three are independent, yet
existing designs couple them, as summarized in
Figure~\ref{fig:design-space}, producing three common trade-offs:

\begin{figure}[t]
    \centering
    \includegraphics[width=\textwidth]{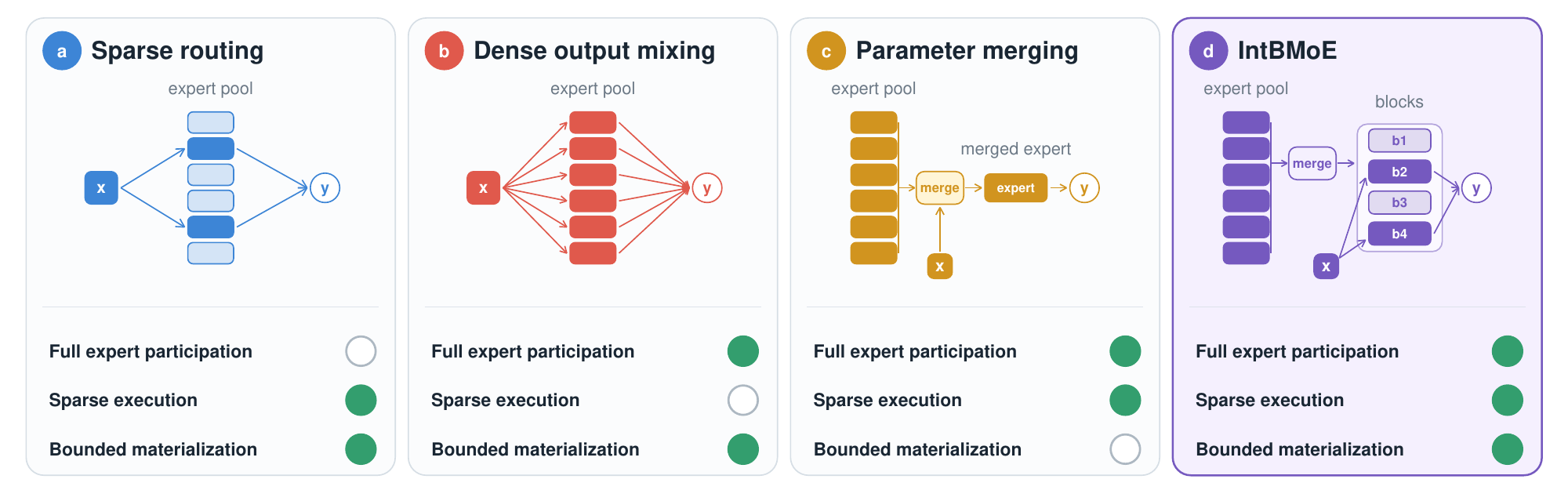}
    \caption{Comparison of MoE design strategies. Filled circles indicate that
    a strategy satisfies the corresponding property, while hollow circles
    indicate otherwise.}
    \label{fig:design-space}
\end{figure}

\begin{itemize}
    \item \textbf{Sparse execution limits participation.} Sparse-routing methods bound execution by
    activating only a few experts for each token. The decision narrows
    participation: non-selected experts neither affect the token's output nor
    receive a learning signal from it.
    \item \textbf{Full participation requires dense execution.} Dense
    output-mixing methods combine the outputs of all experts for each token.
    Because every expert must be evaluated, execution cost grows with the
    number of experts.
    \item \textbf{Single-expert execution increases materialization.} Parameter-merging
    methods combine the full expert pool into one composed expert and execute it
    once. Every distinct routing decision, however, needs its own copy of those
    composed weights, so the number of expert-sized parameter sets built at run
    time grows with the number of routing units.
\end{itemize}

Taken together, these trade-offs leave a central question: \textbf{Can every
token benefit from the full expert pool without requiring dense execution or
unbounded parameter materialization?} Existing MoE formulations cannot satisfy
all three requirements because they couple the construction of expert
transformations with their execution on tokens. To break this coupling, we
propose IntBMoE, a block-conditioned MoE architecture that separates the two
stages. The full expert pool first constructs a bounded set of reusable
transformations, after which each token independently selects and executes only
a few of them. IntBMoE thereby achieves pool-wide expert participation, sparse
execution, and bounded parameter materialization.

Our main contributions are summarized as follows:
\begin{itemize}
    \item \textbf{Decoupling participation, execution, and materialization.} We propose IntBMoE to set the three quantities independently. A small codebook of learned embeddings defines the blocks a module can use, and a shared hypernetwork builds each one layer by layer, merging that layer's expert bases into a single composed expert. A router then sends each token to only a few of these blocks. Participation is therefore pool-wide, execution stays sparse, and materialization is bounded by the codebook.
    \item \textbf{Expressive expert composition.} We introduce Dual-Path Residual Gating (DPRG), which merges each block's expert bases twice, into a value path and a gate path whose product is nonlinear in those bases. DPRG therefore adds expressiveness without enlarging the expert pool.
    \item \textbf{Visual evaluation and cross-domain generalization.} We compare
    IntBMoE with representative sparse and dense MoE baselines on ImageNet-1K
    and observe consistent improvements. Results on language modeling and
    sequential recommendation further show that the same architecture is
    effective across modalities and application domains. IntBMoE
    has also been fully deployed in AMap's generative recommendation system,
    serving hundreds of millions of users under a strict $60\,\mathrm{ms}$
    latency budget and delivering a $2.4\%$ relative UVCTR improvement in
    large-scale online A/B testing.
\end{itemize}

\section{Related Work}

\subsection{Sparse Mixture-of-Experts}

Sparsely gated MoEs activate only a subset of experts for each
token~\cite{shazeer2017outrageously}. GShard~\cite{lepikhin2020gshard}
combines Top-2 expert routing with automatic sharding for large-scale
Transformer training, while Switch Transformer~\cite{fedus2022switch}
simplifies the routing rule to Top-1 selection. V-MoE~\cite{riquelme2021scaling}
extends token-choice sparse routing to Vision Transformers by routing image
patch tokens to a small number of experts. Expert Choice~\cite{zhou2022mixture}
reverses the assignment direction. Each expert selects a fixed-capacity set of
tokens, balancing expert workloads while allowing a variable number of
assignments per token. DeepSeekMoE~\cite{dai2024deepseekmoe}
improves expert organization through fine-grained expert segmentation and
shared-expert isolation. DeepSeek-V3~\cite{liu2024deepseek} retains this
fine-grained architecture and introduces an auxiliary-loss-free routing bias
to improve load balance.

Subsequent work modifies expert structure and routing while preserving sparse
execution. D$^2$-MoE~\cite{gu2025delta} decomposes pretrained experts into a
shared base and compressed expert-specific deltas, after which sparse routing
activates only the selected deltas. ReLU-routing ReMoE~\cite{wang2024remoe}
replaces discontinuous Top-$k$ selection with continuous ReLU gates and
regularizes their sparsity and load balance. LapSum SoftMoE~\cite{zasada2026softmoe} instead uses a truncated soft Top-$k$ relaxation
and learns how to allocate an overall expert-computation budget across layers.
Dense2MoE~\cite{zheng2025dense2moe} extends sparse selection to model depth.
Its Mixture of Blocks executes only a subset of existing Transformer blocks.

Recent adaptive sparse MoEs relax fixed choices for the number of experts
maintained per layer and activated per token. DynMoE~\cite{guo2025dynamic}
adjusts the expert pool based on token--expert routing coverage and uses Top-any routing.
MASS~\cite{park2026many} expands the expert pool using
gradient-based semantic drift detection and uses Top-$p$ routing.

Together, these methods improve scalability, routing efficiency, and expert
organization. However, expert participation remains coupled to execution.
Each token can benefit only from the experts selected and evaluated for it.

\subsection{Dense Output Mixing}

Dense output-mixing methods evaluate all experts and combine their outputs with
learned routing weights. MMoE~\cite{ma2018modeling} shares an expert pool across
tasks and learns a task-specific gate to combine the expert outputs.
PLE~\cite{tang2020progressive} stacks multiple extraction layers containing
shared and task-specific experts, progressively separating shared knowledge
from task-specific information. Both allow every expert in the relevant pool
to contribute to an output, but doing so requires computing every expert’s
output, causing the execution cost to grow with the number of experts.

Soft MoE~\cite{puigcerver2024sparse} provides a distinct slot-based variant.
It softly aggregates input tokens into a fixed set of slots, processes each
slot with its assigned expert, and maps the processed slots back to individual
tokens. Because each slot mixes
all input tokens, the original formulation does not preserve causality and is
not directly applicable to autoregressive prediction.
$\mu$MoE~\cite{oldfield2024multilinear} instead takes a factorized approach.
It represents expert weights as a tensor and computes their mixture using CP
or Tensor Ring factorization. This avoids materializing the full tensor and
evaluating experts separately.

\subsection{Parameter Merging}

Parameter-merging methods achieve full expert participation in a different way. SMEAR~\cite{muqeeth2023soft} constructs one composite expert by taking a
routing-weighted average of all experts and
then executes the merged expert on the input. In its example-level form, all
tokens in a sequence share the same merged expert. This prevents the composition from adapting to individual tokens. A composition derived from the complete sequence also uses future information, so it cannot be applied directly to causal prediction. In its token-level form, SMEAR produces a separate composition for each token. This enables token-specific adaptation but requires a full-pool parameter merge per token, substantially increasing materialization cost. Lory~\cite{zhong2024lory} reduces the cost of this operation through causal segment-level routing. A composition derived from the preceding segment is reused by all tokens in the current
segment. During generation, a prompt-conditioned composition is reused within
the request. This reduces the frequency of parameter merging, but all tokens
within a segment share the same composition, limiting token-level adaptation.

DSFNet~\cite{yu2025dsfnet} performs input-conditioned parameter merging to
construct scenario-specific network parameters, using gates to linearly
combine parameter sets from disentangled factor-scenario branches. In
contrast, IntBMoE preconstructs a finite set of input-independent blocks and
adapts to each token through sparse block routing.

\subsection{Hypernetworks and Parameter Generation}

Hypernetworks~\cite{ha2017hypernetworks} generate the parameters of a target network from learned or dynamically produced conditioning embeddings. Directly generating full weight matrices can be memory-intensive. When the conditioning signal is input-dependent, the parameters must also be regenerated as the signal changes.

HyperMoE~\cite{zhao2024hypermoe} applies this idea to sparse MoE. It encodes information associated with a token's unselected experts and generates a HyperExpert for that token. The HyperExpert is executed alongside the selected experts, providing an additional token-conditioned path while the unselected experts remain inactive. IntBMoE conditions its hypernetwork on a finite codebook of learned, input-independent block embeddings. The hypernetwork generates compact coefficients that combine a shared pool of expert bases, rather than directly generating full weight matrices. Because the conditioning set is finite and input-independent, the composed blocks can be precomputed and reused across routing decisions. IntBMoE can therefore be viewed as a basis-constrained hypernetwork with bounded parameter materialization.

\section{Preliminaries}

Let $\mathbf x_t\in\mathbb R^d$ denote the representation of input element
$t$, and let $f_e(\cdot;\boldsymbol\theta_e)$ denote expert $e$ with parameters
$\boldsymbol\theta_e$. A conventional sparse MoE maintains $E$ experts and
uses a router to select $k$ of them:
\begin{equation}
    \mathbf r_t=g_{\mathrm{router}}(\mathbf x_t),
    \qquad \mathbf p_t=\operatorname{softmax}(\mathbf r_t),
    \qquad
    \mathbf y_t^{\mathrm{sparse}}=
    \sum_{e\in\TopK(\mathbf r_t,k)}
    p_{t,e}f_e(\mathbf x_t;\boldsymbol\theta_e).
\end{equation}
Here, $g_{\mathrm{router}}:\mathbb R^d\rightarrow\mathbb R^E$ is the
routing network, $\mathbf r_t$ contains its scores for the $E$ experts, and
$p_{t,e}$ is the routing probability assigned to expert $e$.
$\TopK(\mathbf r_t,k)$ returns the indices of the $k$ highest-scoring experts.
Sparse routing evaluates only $k$ expert networks for the token.

A direct dense output mixture instead aggregates all expert outputs,
\begin{equation}
    \mathbf y_t^{\mathrm{dense}}=
    \sum_{e=1}^{E}p_{t,e}f_e(\mathbf x_t;\boldsymbol\theta_e).
\end{equation}
All $E$ experts can therefore affect the token, but every expert must be executed. Slot-based variants change the unit of expert computation
from individual tokens to learned token mixtures, but still process the full
set of expert-associated slots.

Parameter-merging methods obtain full expert participation without separately
executing every expert. Let $u$ denote a routing unit, which may be a token, segment, or sequence. We use $a_{u,e}$ for the coefficient assigned to expert $e$. These methods
construct
\begin{equation}
    \bar{\boldsymbol\theta}_u=
    \sum_{e=1}^{E}a_{u,e}\boldsymbol\theta_e,
    \qquad
    \mathbf y_t^{\mathrm{merge}}=
    f(\mathbf x_t;\bar{\boldsymbol\theta}_u).
\end{equation}
The token is processed by one composite expert, but each routing unit requires its own expert-sized parameter set, constructed by combining all experts.
If a layer contains $G$ distinct routing units, parameter synthesis costs
$O(GE|\boldsymbol\theta|)$, where $|\boldsymbol\theta|$ denotes the size of one
expert. Here, \emph{materialization} refers to constructing these derived
parameter sets at runtime.

\section{Method}
\subsection{Architecture Overview}

\begin{figure*}[t]
    \centering
    \includegraphics[width=\textwidth]{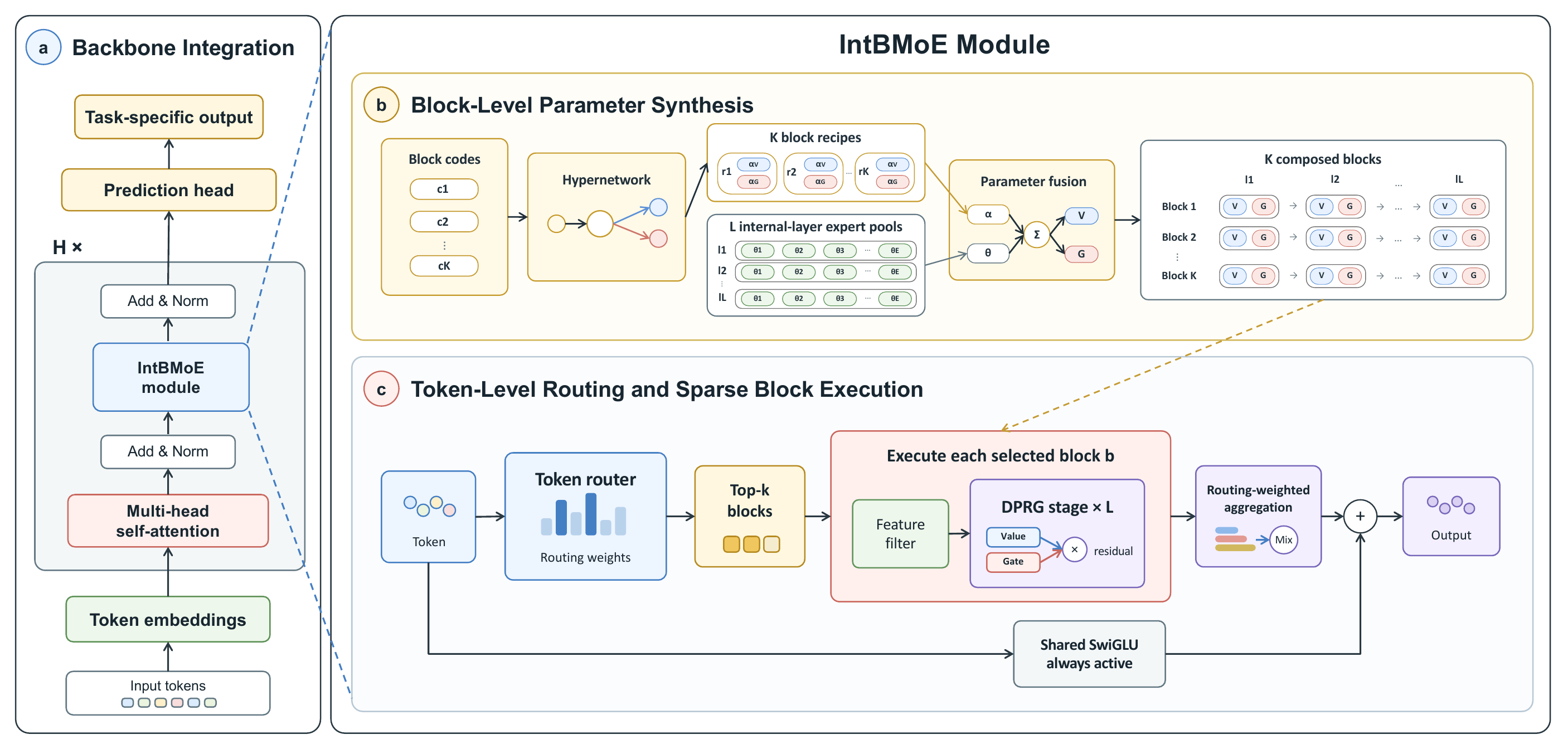}
    \caption{Overview of IntBMoE. The left panel shows independent IntBMoE
    modules integrated into a multi-layer Transformer backbone. The right panel
    shows block-level expert parameter synthesis and token-level Top-$k$ block
    routing and execution.}

    \label{fig:overview}
\end{figure*}

IntBMoE integrates into a Transformer backbone~\cite{vaswani2017attention} by
replacing its FFN sublayer. Each IntBMoE module composes reusable multi-layer
blocks from shared expert pools and sparsely routes each token to a few blocks,
as illustrated in Figure~\ref{fig:overview}. At the block level, $K$ learned codebook embeddings
produce composition coefficients. These coefficients combine
layer-specific expert pools into $K$ token-independent $L$-layer blocks. At the token level, a
router independently selects the Top-$k$ blocks for each token. Each
selected block applies block-conditioned feature filtering and processes the
token sequentially through its $L$ composed experts. The selected outputs are
weighted by their routing probabilities and combined with the output of an
always-active shared SwiGLU expert. The following subsections describe these
components in detail.

\subsection{Block-Level Parameter Synthesis}

Block synthesis starts from $K$ learned codebook embeddings, one per candidate
block. A shared hypernetwork maps each embedding to value and gate composition
coefficients, which combine the layer-wise expert pools into $K$ reusable
multi-layer blocks. The token router then selects among these blocks, as
described in the next subsection.

Concretely, each IntBMoE module maintains a codebook of $K$ learned embeddings,
\begin{equation}
    \mathcal C=\{\mathbf c_b\in\mathbb R^{d_c}\}_{b=1}^{K}.
\end{equation}
Each embedding $\mathbf c_b$ identifies one $L$-layer block and is used to
synthesize its parameters. The codebook therefore defines the $K$ blocks
available to the token router.

We use column vectors throughout; linear maps act by left multiplication.
Let $d_0=d_L=d$, with
$d_\ell$ denoting the output dimension of internal layer $\ell$. For each
$\ell\in\{1,\ldots,L\}$, the module maintains a layer-specific pool of $E$
expert bases shared across the $K$ blocks,

\begin{equation}
    \mathcal P^{(\ell)}=
    \{(W_e^{(\ell)},\mathbf b_e^{(\ell)})\}_{e=1}^{E},
    \quad
    W_e^{(\ell)}\in\mathbb R^{d_\ell\times d_{\ell-1}},
    \quad
    \mathbf b_e^{(\ell)}\in\mathbb R^{d_\ell}.
\end{equation}

Here, $\mathcal P^{(\ell)}$ denotes the expert pool at internal layer $\ell$,
and $(W_e^{(\ell)},\mathbf b_e^{(\ell)})$ is its $e$-th expert.
These routed experts are distinct from the always-active shared expert introduced
later. The pools are independent across internal layers and backbone layers.

The block hypernetwork $h_\phi$ is shared across the $K$ codebook entries within
one IntBMoE module. It takes only the learned block embedding $\mathbf c_b$ as input and
uses a linear--LayerNorm--ReLU trunk followed by two linear output heads. Its
hidden representation is
\begin{equation}
    \mathbf q_b=\operatorname{ReLU}\!\left(
    \operatorname{LN}(A\mathbf c_b+\mathbf a)
    \right),
\end{equation}
where $A\in\mathbb R^{d_h\times d_c}$ and
$\mathbf a\in\mathbb R^{d_h}$ are the trainable weight matrix and bias of the
input projection. The resulting representation satisfies
$\mathbf q_b\in\mathbb R^{d_h}$. Two linear output heads
map $\mathbf q_b$ to the value and gate composition coefficients
$\boldsymbol\alpha_b^v$ and $\boldsymbol\alpha_b^g$, respectively:
\begin{equation}
    \boldsymbol\alpha_b^v=A_v\mathbf q_b+\mathbf a_v,
    \qquad
    \boldsymbol\alpha_b^g=A_g\mathbf q_b+\mathbf a_g,
    \qquad
    \boldsymbol\alpha_b^v,\boldsymbol\alpha_b^g\in\mathbb R^E.
    \label{eq:hypernetwork}
\end{equation}
Here, $A_v,A_g\in\mathbb R^{E\times d_h}$ and
$\mathbf a_v,\mathbf a_g\in\mathbb R^E$ are the trainable weights and biases
of the two output heads.

The coefficients are not normalized by softmax or sigmoid. They may be
negative and need not sum to one, allowing composition over the linear span of
the expert bases rather than restricting it to their convex hull. We instead
apply a variance-preserving factor of $1/\sqrt{E}$ when composing the expert
bases, keeping the scale of the composed parameters approximately stable as
the expert pool grows. For path $p\in\{v,g\}$, the composed parameters at every
internal layer are

\begin{equation}
    W_{b,p}^{(\ell)}=\frac{1}{\sqrt{E}}\sum_{e=1}^{E}\alpha^{p}_{b,e}W_e^{(\ell)},
    \qquad
    \mathbf b_{b,p}^{(\ell)}=\frac{1}{\sqrt{E}}\sum_{e=1}^{E}\alpha^{p}_{b,e}\mathbf b_e^{(\ell)}.
    \label{eq:block-compose}
\end{equation}

Thus, the hypernetwork produces one recipe
$(\boldsymbol\alpha_b^v,\boldsymbol\alpha_b^g)$ for block $b$. At each internal
layer, this recipe combines that layer's expert pool into separate value and
gate parameter sets. Repeating this process over all $L$ internal layers
constructs the complete $L$-layer block.
Because $h_\phi$ does not depend on token representations, the $K$ composed
blocks can be precomputed and shared across all tokens.

\subsection{Token-Level Block Routing}

Routing is performed independently for each token. For token $t$, the router
uses its representation $\mathbf x_t$ to produce one score for each block and
selects the Top-$k$ blocks:
\begin{equation}
    \mathbf r_t=g_{\mathrm{block\text{-}router}}(\mathbf x_t),
    \qquad
    \mathcal B_t=\TopK(\mathbf r_t,k),
\end{equation}
where $\mathcal B_t$ is the set of indices of the $k$ blocks selected for
token $t$. The block router
$g_{\mathrm{block\text{-}router}}:\mathbb R^d\rightarrow\mathbb R^K$
outputs a score vector $\mathbf r_t\in\mathbb R^K$ and is implemented as a
two-layer ReLU MLP by default. The routing weight of a selected block $b$ is
\begin{equation}
    \pi_{t,b}=\softmax(\mathbf r_t)_b,
    \qquad b\in\mathcal B_t.
\end{equation}

\subsection{Block-Conditioned Feature Filtering}

Before entering a selected block, the token representation $\mathbf x_t$ is
filtered using that block's codebook embedding $\mathbf c_b$:
\begin{equation}
    \mathbf m_{t,b}=\operatorname{sigmoid}
    \left(\mathbf W_f[\mathbf x_t\mathbin\Vert\mathbf c_b]+\mathbf b_f\right),
    \qquad
    \mathbf z^{(0)}_{t,b}=\mathbf x_t\odot\mathbf m_{t,b}.
\end{equation}
Here, $[\cdot\mathbin\Vert\cdot]$ denotes vertical concatenation of column vectors,
$\mathbf W_f\in\mathbb R^{d\times(d+d_c)}$ and
$\mathbf b_f\in\mathbb R^d$ are the learnable parameters of a feature-filtering
layer shared across blocks, and $\mathbf m_{t,b}\in(0,1)^d$ is a soft
feature-wise mask. The filtered representation
$\mathbf z^{(0)}_{t,b}\in\mathbb R^d$ is the input to the first internal layer
of block $b$ for token $t$.
This gives different blocks distinct views of the same token before their
composed transformations are applied.

\subsection{Dual-Path Residual Gating}

Although each parameter path in~\eqref{eq:block-compose} is composed
linearly from the expert bases, DPRG introduces a nonlinear interaction between
two independently composed paths. It couples the value and gate paths through
residual multiplicative modulation.

For internal layer $\ell$ and input $\mathbf z^{(\ell-1)}_{t,b}$, the DPRG
transformation is
\begin{align}
    \mathbf v^{(\ell)}_{t,b} &=
    W_{b,v}^{(\ell)}\mathbf z^{(\ell-1)}_{t,b}+\mathbf b_{b,v}^{(\ell)},\\
    \mathbf g^{(\ell)}_{t,b} &=
    \operatorname{RMSNorm}\!\left(
    W_{b,g}^{(\ell)}\mathbf z^{(\ell-1)}_{t,b}+\mathbf b_{b,g}^{(\ell)}
    \right),\\
    \widetilde{\mathbf z}^{(\ell)}_{t,b} &=
    \mathbf v^{(\ell)}_{t,b}\odot
    \left(\mathbf 1+\lambda\,\operatorname{SiLU}(\mathbf g^{(\ell)}_{t,b})\right).
    \label{eq:dprg}
\end{align}
LayerNorm is applied between consecutive internal layers:
\begin{equation}
    \mathbf z^{(\ell)}_{t,b} =
    \begin{cases}
        \operatorname{LN}(\widetilde{\mathbf z}^{(\ell)}_{t,b}),
        & \ell<L,\\
        \widetilde{\mathbf z}^{(\ell)}_{t,b},
        & \ell=L.
    \end{cases}
\end{equation}

Thus, each layer processes the token state produced by the preceding layer, and
$F_b(\mathbf x_t)=\mathbf z^{(L)}_{t,b}$ is the final output of block $b$. The
learnable residual scale $\lambda$ is shared across the internal layers of an
IntBMoE module. DPRG increases the expressiveness of each composed block by coupling
two compositions of the same expert pool, while adding only a constant factor
to its parameter-synthesis and execution costs.

\subsection{Output Aggregation and Shared Expert}

Selected block outputs are aggregated using routing probabilities:
\begin{equation}
    \mathbf y^{\mathrm{route}}_t=
    \sum_{b\in\mathcal B_t}\pi_{t,b}F_b(\mathbf x_t).
\end{equation}
We add an always-active shared SwiGLU expert $S$ to model components that need
not be differentiated by block routing:
\begin{equation}
    \mathbf y_t=\mathbf y^{\mathrm{route}}_t+S(\mathbf x_t).
\end{equation}

The shared path captures common information, allowing the routed blocks to
focus on transformations that benefit from token-dependent selection.

\subsection{Complexity and Inference Caching}

Consider an IntBMoE module processing $T$ valid tokens. Let each block contain
$L$ layers, where layer $\ell$ maps
$d_{\ell-1}$ to $d_\ell$, and define
$D=\sum_{\ell=1}^{L}d_{\ell-1}d_\ell$ as the total matrix size of one
multi-layer expert basis. Composing all $K$ blocks from $E$ bases costs
$O(KED)$, and executing the $k$ selected blocks for all tokens costs
$O(TkD)$. Ignoring lower-order routing operations and the constant factor from
the two DPRG paths, the total uncached composition-and-routed-execution cost is
$O(KED+TkD)$.
The corresponding amortized per-token cost is $O(KED/T+kD)$. The composition
term remains linear in $E$, but it is incurred once for $K$ reusable blocks
rather than once per token.

Block composition
depends only on the learned block embeddings, hypernetwork, and expert bases.
Once the model parameters are fixed for inference, all composed block
parameters can be constructed once and cached, removing the $O(KED)$
composition term from request-time computation. The dominant routed-block cost
is then $O(TkD)$, in addition to the router, feature filter, and shared expert.
Consequently, for fixed $K$ and $k$, the request-time computation of cached
IntBMoE does not increase with the expert-pool size $E$.

\section{Visual Experiments}
\label{sec:visual-experiments}

\subsection{Experimental Setup}
\label{sec:visual-setup}

ImageNet-1K~\cite{russakovsky2015imagenet} is our primary benchmark. It
contains 1.28 million training images and 50,000 validation images from 1,000
classes. All methods use an eight-layer DeiT-Tiny-style
backbone~\cite{touvron2021training} and are trained from scratch on the official
training split. We report the mean Top-1 and Top-5 validation accuracy over
three random seeds.

We compare IntBMoE with the dense backbone and two families of MoE methods.
Sparse-routing baselines include Switch
Transformer~\cite{fedus2022switch}, DeepSeek-V3 MoE~\cite{liu2024deepseek},
ReMoE~\cite{wang2024remoe}, V-MoE~\cite{riquelme2021scaling}, Expert
Choice~\cite{zhou2022mixture}, DynMoE~\cite{guo2025dynamic}, and
MASS~\cite{park2026many}. Dense-participation
baselines include Soft MoE~\cite{puigcerver2024sparse},
SMEAR~\cite{muqeeth2023soft}, Lory~\cite{zhong2024lory}, and
$\mu$MoE~\cite{oldfield2024multilinear}. Complete experimental configurations
and implementation details are provided in
Appendix~\ref{app:experimental-configurations}.

\subsection{Experimental Results}

\begin{table}[t]
\centering
\small
\caption{Main results on ImageNet-1K. $\uparrow$ indicates that higher values
are better.
Best results are highlighted with a \bestlabel{blue background}, and
second-best results are shown in \second{blue}. Params denotes the total number
of trainable parameters in the model; Activated Params denotes the number of
trainable parameters activated as a single patch token passes through the
entire model. FLOPs denotes the inference computational cost per image, measured with a batch size of 1.}
\label{tab:main-vision}
\vspace{0.5em}
\setlength{\tabcolsep}{5pt}
\begin{tabular*}{0.96\textwidth}{@{\extracolsep{\fill}}lccccc@{}}
\toprule
Method & ACC@1$\uparrow$ & ACC@5$\uparrow$ & Params (M) &
Activated (M) & FLOPs (G)\\
\midrule
Dense & 0.6640 & 0.8769 & 3.938 & 3.938 & 1.453\\
\midrule
Switch Transformer & 0.7004 & 0.8937 & 24.070 & 3.951 & 1.458\\
DeepSeek-V3 MoE & 0.7078 & 0.8969 & 24.004 & 6.306 & 2.382\\
ReMoE & 0.6971 & 0.8946 & 24.070 & 4.574 & 1.458\\
V-MoE & 0.7167 & 0.9030 & 24.070 & 5.135 & 1.920\\
Expert Choice & 0.7170 & 0.9030 & 24.070 & 5.135 & 1.958\\
DynMoE & 0.6975 & 0.8962 & 24.070 & 10.306 & 4.170\\
MASS & 0.7008 & 0.8966 & 24.070 & 8.827 & 3.601\\
\midrule
Soft MoE & 0.7122 & 0.8996 & 24.201 & 23.017 & 1.656\\
SMEAR & \second{0.7178} & \second{0.9033} & 24.070 & 22.887 & 1.493\\
Lory & 0.6956 & 0.8892 & 24.005 & 22.822 & 2.003\\
$\mu$MoE (CP) & 0.7012 & 0.8938 & 24.070 & 22.887 & 9.120\\
$\mu$MoE (TR) & 0.6962 & 0.8929 & 24.065 & 22.882 & 9.151\\
\midrule
IntBMoE & \best{0.7376} & \best{0.9148} & 24.295 & 23.111 & 4.063\\
IntBMoE (cached) & -- & -- & -- & -- & 3.457\\
\bottomrule
\end{tabular*}
\end{table}

\paragraph{Main results.}
Table~\ref{tab:main-vision} presents the primary comparison. IntBMoE achieves
73.76\% Top-1 and 91.48\% Top-5 accuracy. Relative to the dense backbone, these
scores represent gains of 7.36 and 3.79 percentage points, respectively.
IntBMoE also outperforms all sparse-routing and dense-participation MoE
baselines, exceeding the strongest competitor, SMEAR, by 1.98 Top-1 percentage
points and 1.15 Top-5 percentage points.

All MoE methods use comparable total parameter budgets of approximately 24M.
IntBMoE activates 23.111M parameters because the full expert pool participates
in constructing its reusable blocks. Despite full expert participation, each
token executes only its selected blocks. The total inference cost is 4.063
GFLOPs per image without caching and 3.457 GFLOPs per image with caching,
measured with a batch size of 1.

\paragraph{Ablation studies.}

To assess the contribution of each component, Table~\ref{tab:ablation-vision}
reports ablations on ImageNet-1K. The \emph{1-Layer} variant replaces each
two-layer block with a single layer and enlarges its expert pool to preserve
the total number of
expert parameters. \emph{Fixed $\lambda$} makes $\lambda$ non-learnable and
sets it to 1. \emph{w/o Gate} removes the composed gate path and multiplicative
modulation, reducing~\eqref{eq:dprg} to
$\widetilde{\mathbf z}^{(\ell)}_{t,b}=\mathbf v^{(\ell)}_{t,b}$.
\emph{w/o Shared} removes the shared SwiGLU expert, while \emph{w/o Filter}
bypasses block-conditioned feature filtering by setting
$\mathbf z^{(0)}_{t,b}=\mathbf x_t$. Finally, \emph{Softmax Coeff.} replaces
our unconstrained, variance-scaled coefficient formulation with separately
softmax-normalized value and gate coefficients.

\begin{table}[t]
\centering
\small
\caption{Component ablations on ImageNet-1K.}
\label{tab:ablation-vision}
\setlength{\tabcolsep}{3.5pt}
\renewcommand{\arraystretch}{1.15}
\begin{tabular*}{\textwidth}{@{\extracolsep{\fill}}lccccccc@{}}
\toprule
Metric & \textbf{Full} & 1-Layer & Fixed $\lambda$ & w/o Gate &
w/o Shared & w/o Filter & Softmax Coeff.\\
\midrule
Top-1$\uparrow$ & \textbf{0.7376} & 0.6878 & 0.7244 & 0.6804 &
0.7332 & 0.7372 & 0.7263\\
Top-5$\uparrow$ & \textbf{0.9148} & 0.8910 & 0.9078 & 0.8831 &
0.9108 & 0.9132 & 0.9072\\
\bottomrule
\end{tabular*}
\end{table}

Removing the gate path causes the largest performance drop, followed by
collapsing each two-layer block into a parameter-matched single layer. The
shared expert and feature filter provide smaller but consistent gains. Both
fixing $\lambda$ and softmax-normalizing the composition coefficients
underperform the full formulation.

\subsection{Sensitivity to Architectural Hyperparameters}

\begin{figure*}[t]
    \centering
    \includegraphics[width=\textwidth]{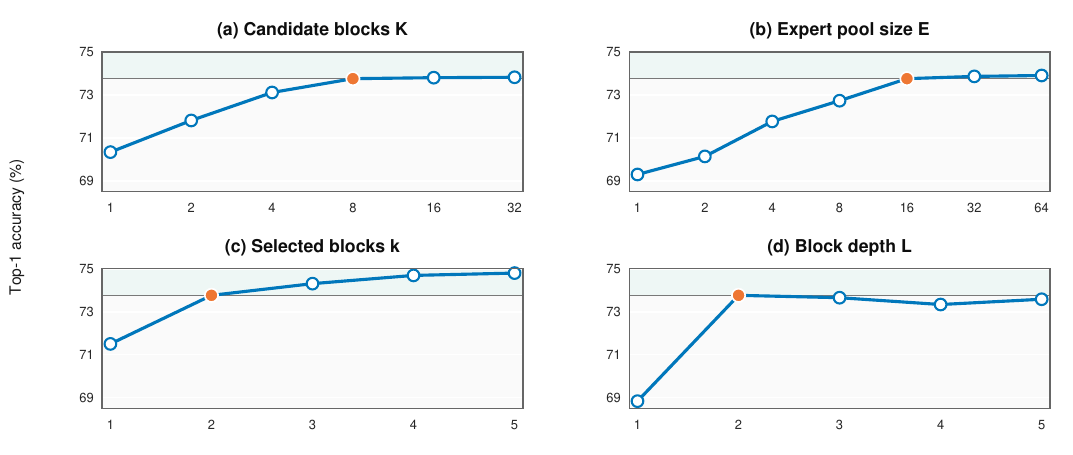}
    \caption{Sensitivity of IntBMoE to architectural hyperparameters on
    ImageNet-1K. Each panel reports Top-1 accuracy while varying one
    hyperparameter; higher values are better. All panels use the same vertical
    scale.
    Orange points and gray horizontal lines mark the default configuration
    $(K,E,k,L)=(8,16,2,2)$ and its accuracy, respectively.}
    \label{fig:sensitivity}
\end{figure*}

We vary the codebook size $K$, the number of selected blocks $k$, the
expert-pool size $E$ at each internal layer, and the block depth $L$ on
ImageNet-1K. We change one hyperparameter at a time from the default
configuration $(K,E,k,L)=(8,16,2,2)$. For the $K=1$ setting, we set $k=1$.
Figure~\ref{fig:sensitivity} reports the resulting Top-1 accuracies.

Increasing either $K$ or $E$ improves accuracy, but further gains are small
beyond their default values. Expanding $K$ from 8 to 32 increases Top-1
accuracy by only 0.07 percentage points, and expanding $E$ from 16 to 64
increases it by only 0.15 percentage points. Increasing $k$
consistently improves Top-1 accuracy. We use $k=2$ to obtain a substantial gain
over $k=1$ while limiting per-token computation. For block depth, $L=2$
achieves the highest Top-1 accuracy.

\subsection{Effective Full-Pool Expert Participation}
Although every block is constructed from all expert bases, this design alone
does not show that every expert makes a meaningful contribution to model
performance. We therefore evaluate the contribution of each expert basis on
ImageNet-1K by removing it from all block compositions in one MoE layer at a
time. We examine all 16 expert bases in Layers 0, 2, 4, and 6, producing 64
removal settings in total, and evaluate the original checkpoint without
retraining. Removing an expert may reduce the
magnitude of the composed weights simply because fewer expert bases remain.
To isolate the contribution of the removed expert, we compensate for this
scale change by replacing $1/\sqrt{E}$ with $1/\sqrt{E-1}$.

\begin{figure*}[t]
    \centering
    \includegraphics[width=0.94\textwidth]{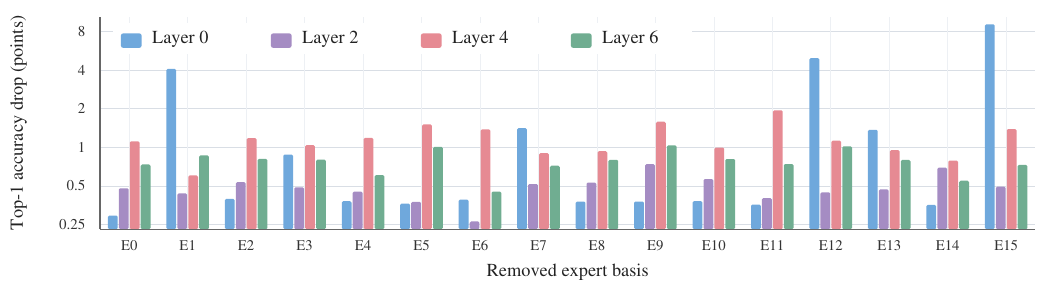}
    \caption{Top-1 accuracy drop on ImageNet-1K after removing individual
    expert bases from one MoE layer at a time. The vertical axis uses a
    logarithmic scale.}
    \label{fig:expert-removal}
\end{figure*}

Figure~\ref{fig:expert-removal} shows that removing any of the examined expert bases
reduces Top-1 accuracy. The mean drops at Layers 0, 2, 4, and 6 are 1.60, 0.49,
1.17, and 0.78 percentage points, respectively. Contributions are highly
uneven in Layer 0, where removing E1, E12, or E15 lowers accuracy by 4.10,
4.99, and 9.15 percentage points, while the effects of most other bases are
much smaller. The removal effects become more balanced in later layers,
ranging from 0.26 to 0.74 percentage points in Layer 2, 0.60 to 1.95
percentage points in Layer 4, and 0.45 to 1.04 percentage points in Layer 6.
This suggests that the earliest MoE layer relies strongly on a few expert
bases, whereas later layers distribute useful contributions more evenly
across their pools. Even the smallest decrease is 0.26 percentage points,
indicating that every expert basis contributes to the model.
Together, the results support the conclusion that the model makes effective
use of the full expert pool throughout its MoE layers.

\subsection{Block Routing and Block-Specific Composition Coefficients}

\begin{figure*}[t]
    \centering
    \includegraphics[width=\textwidth]{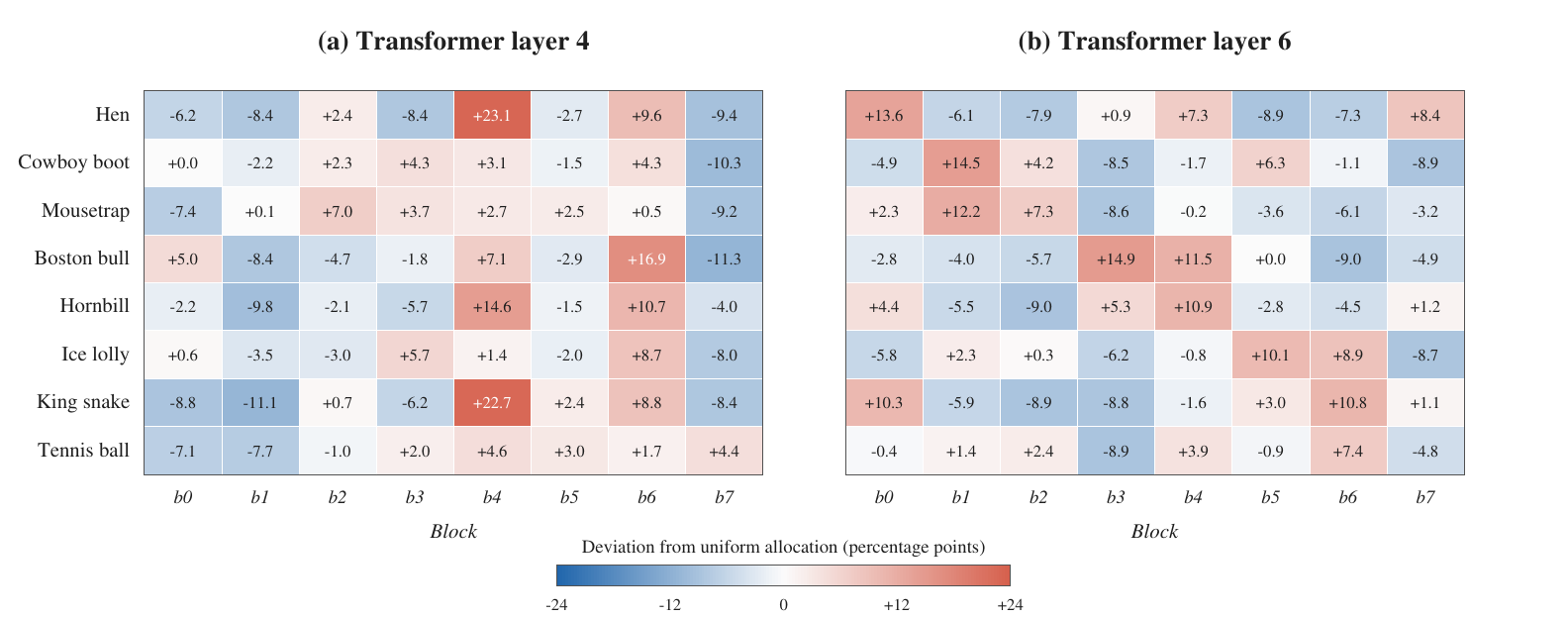}
    \caption{Class-conditioned block routing for the same eight representative
    ImageNet-1K classes at Transformer layers 4 and 6. Each cell shows the
    token-assignment proportion for a class and block minus the uniform
    allocation of 12.5\%, in percentage points. Positive (red) and negative
    (blue) values indicate above- and below-uniform use of a block,
    respectively. Each Top-$k$ selection is counted as one token--block
    assignment.}
    \label{fig:block-routing}
\end{figure*}

\paragraph{Class-conditioned block allocation.}
On ImageNet-1K, we compute the proportion of patch-token routing assignments
received by each block within each ground-truth class. Figure~\ref{fig:block-routing}
shows deviations from the uniform 12.5\% allocation for the same eight
representative classes at layers 4 and 6.

Different image classes favor different blocks within the same layer, and
these preferences often change with depth. For example, at layer 4, hen favors
$b_4$, whereas Boston bull favors $b_6$. At layer 6, their preferred blocks
shift to $b_0$ and $b_3$, respectively. These patterns show that routing is
both class-conditioned and depth-dependent.

\begin{figure*}[t]
    \centering
    \includegraphics[width=\textwidth]{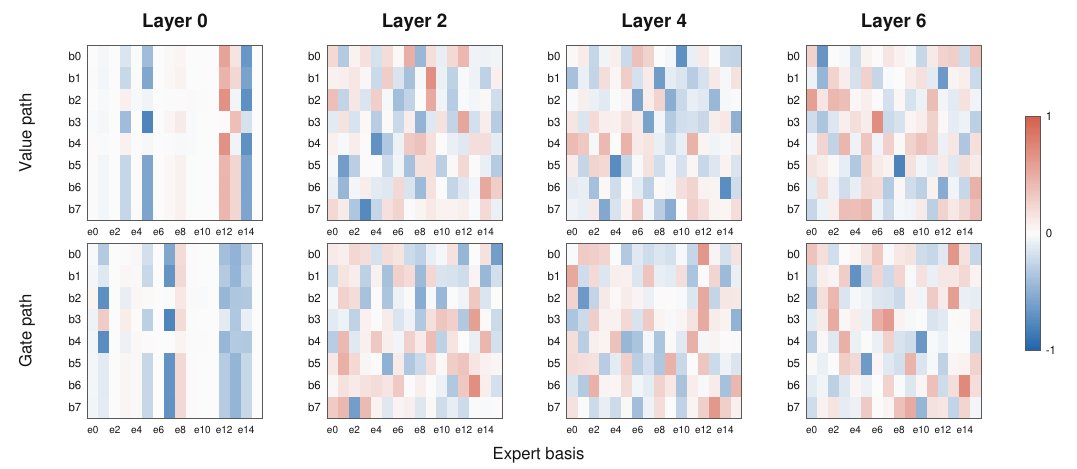}
    \caption{Expert-composition coefficients learned by the eight blocks in
    Transformer layers 0, 2, 4, and 6 on ImageNet-1K, ordered from shallow to
    deep along the backbone. Rows represent blocks and columns represent expert
    bases. Each block's coefficient vector is
    $\ell_2$-normalized only for visualization. Red and blue denote positive and
    negative coefficients, respectively.}
    \label{fig:composition-coefficients}
\end{figure*}

\paragraph{Block-specific composition coefficients.}

We visualize the expert-composition coefficients generated for every block on
ImageNet-1K, separately for the value and gate paths. As shown in
Figure~\ref{fig:composition-coefficients}, blocks within the same layer assign
different positive and negative weights to the same expert bases. The patterns
also vary across layers and between the two DPRG paths. Because all blocks in a
layer compose the same ordered expert pools, these differences show that the
codebook entries learn block-specific composition recipes rather than
collapsing to identical fusion coefficients.

The degree of recipe differentiation increases with network depth. For the
value path, the mean pairwise cosine similarity between block recipes decreases
monotonically from 0.796 at layer 0 to 0.079, 0.019, and 0.010 at layers 2, 4,
and 6, respectively. Gate-path similarity follows the same trend, decreasing
from 0.726 to 0.126, 0.083, and 0.043. Thus, the shallowest layer largely shares
an expert-composition pattern across blocks, whereas deeper layers construct
substantially more diverse recipes. This trend is consistent with shallow
visual features being more general and deeper representations benefiting from
finer block-specific expert composition.

\subsection{Inference Cost and Caching}

\begin{figure}[t]
    \centering
    \includegraphics[width=\textwidth]{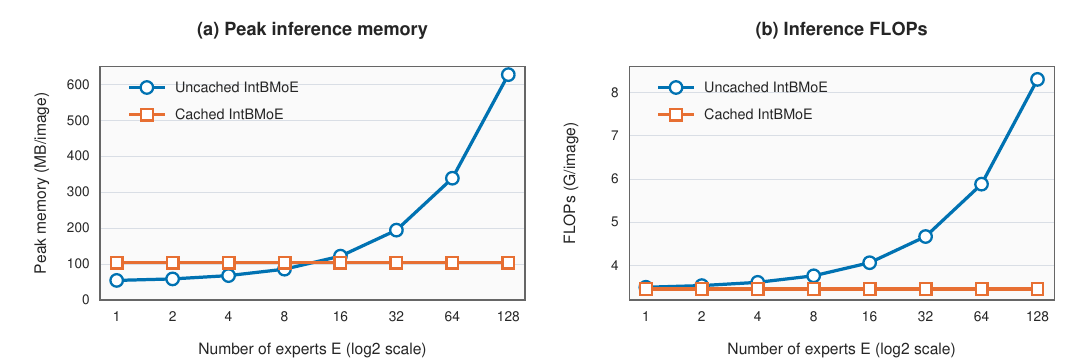}
    \caption{Peak inference memory and inference FLOPs of cached and uncached
    IntBMoE as the number of experts $E$ increases. Both quantities are measured
    for inference on a single image with a batch size of 1 using the ImageNet-1K
    model configuration.
    The horizontal axis uses a $\log_2$ scale.}
    \label{fig:cache-scaling}
\end{figure}

Figure~\ref{fig:cache-scaling} shows how the inference memory and FLOPs of
cached and uncached IntBMoE change as the number of experts increases in the
ImageNet-1K setup. Without caching, peak inference memory increases from
$54.57$ to $627.83$ MB and inference cost
increases from $3.495$ to $8.305$ GFLOPs as the expert-pool size $E$ grows from 1
to 128. With caching, they remain constant at $104.33$ MB and $3.457$ GFLOPs,
respectively. Caching introduces a fixed memory cost
for the precomposed blocks, so it uses more memory for small expert pools, but
becomes more memory-efficient from 16 experts onward. These results show
that caching decouples request-time memory and computation from the size of the
expert pool. During training, all images in a minibatch share the same composed
blocks. The composition is therefore performed once per batch, and its cost is
amortized across the images in that batch.

\section{Generalization to Language Modeling and Recommendation}
\label{sec:generalization}

\paragraph{Tasks and datasets.}
We use MiniPile~\cite{kaddour2023minipile} and
IntTravel~\cite{yan2026inttravel} to test whether the same architecture
generalizes to language modeling and sequential recommendation. MiniPile uses
an 18-layer Llama-style causal Transformer~\cite{touvron2023llama}, while
IntTravel uses a causal Transformer. MiniPile is a 6-GB subset of the
deduplicated Pile. We evaluate causal language modeling on MiniPile using
token-level cross-entropy and perplexity. IntTravel contains 162.8 million
users, 7.3 million POIs, and 4.13 billion interactions. We use its ``Where''
task to predict the destination POI of the next journey and report HR@1, HR@5,
and NDCG@5.

\paragraph{Experimental results.}
Table~\ref{tab:generalization} compares IntBMoE with the dense backbone,
sparse-routing methods, and dense-participation methods on MiniPile and
IntTravel. For these causal prediction tasks, we make Soft MoE causal by
constructing each slot using only the current and preceding tokens. SMEAR uses
token-level composition. Complete model, training,
routing, and baseline configurations are provided in
Appendix~\ref{app:experimental-configurations}.

\begin{table*}[t]
\centering
\small
\caption{Generalization results on MiniPile and IntTravel. $\uparrow$ indicates
that higher values are better, and $\downarrow$ indicates that lower values
are better. Best results per
column are highlighted with a \bestlabel{blue background}, and second-best
results are shown in \second{blue}.}
\label{tab:generalization}
\setlength{\tabcolsep}{5pt}
\begin{tabular*}{\textwidth}{@{\extracolsep{\fill}}lccccc@{}}
\toprule
\multirow{2}{*}{Method} & \multicolumn{2}{c}{MiniPile} &
\multicolumn{3}{c}{IntTravel}\\
\cmidrule(lr){2-3}\cmidrule(lr){4-6}
& Loss$\downarrow$ & PPL$\downarrow$ & HR@1$\uparrow$ & HR@5$\uparrow$ &
NDCG@5$\uparrow$\\
\midrule
Dense & 2.8131 & 16.6621 & 0.6748 & 0.8659 & 0.7785\\
\midrule
Switch Transformer & 2.7156 & 15.1135 & 0.6798 & 0.8667 & 0.7812\\
DeepSeek-V3 MoE & 2.7119 & 15.0580 & 0.6805 & 0.8674 & 0.7818\\
ReMoE & 2.7346 & 15.4034 & 0.6833 & \second{0.8685} & 0.7836\\
DynMoE & 2.7108 & 15.0406 & 0.6818 & 0.8677 & 0.7826\\
MASS & 2.7608 & 15.8128 & 0.6822 & 0.8679 & 0.7829\\
\midrule
Soft MoE (causal) & 2.7112 & 15.0466 & 0.6832 & 0.8683 & 0.7835\\
SMEAR (token-level) & 2.7165 & 15.1277 & 0.6825 & 0.8676 & 0.7828\\
Lory & 2.7519 & 15.6721 & 0.6811 & 0.8677 & 0.7823\\
$\mu$MoE (CP) & \second{2.7101} & \second{15.0306} & 0.6835 & 0.8682 & \second{0.7837}\\
$\mu$MoE (TR) & 2.7171 & 15.1359 & \second{0.6837} & \second{0.8685} & 0.7835\\
\midrule
IntBMoE & \best{2.6802} & \best{14.5878} & \best{0.6852} &
\best{0.8692} & \best{0.7850}\\
\bottomrule
\end{tabular*}
\end{table*}

On MiniPile, IntBMoE achieves a test loss of 2.6802 and a PPL of 14.5878.
Relative to $\mu$MoE (CP), the strongest baseline, IntBMoE reduces PPL by
2.9\%. Relative to the dense backbone, the reduction is 12.4\%. On IntTravel,
IntBMoE achieves the highest mean HR@1, HR@5, and
NDCG@5. These results support generalization to autoregressive language
modeling and sequential recommendation.

\begin{table*}[t]
\centering
\small
\caption{Component ablations on MiniPile and IntTravel.}
\label{tab:ablation-generalization}
\setlength{\tabcolsep}{5pt}
\begin{tabular*}{\textwidth}{@{\extracolsep{\fill}}lccccc@{}}
\toprule
\multirow{2}{*}{Variant} & \multicolumn{2}{c}{MiniPile} &
\multicolumn{3}{c}{IntTravel}\\
\cmidrule(lr){2-3}\cmidrule(lr){4-6}
& Loss$\downarrow$ & PPL$\downarrow$ & HR@1$\uparrow$ & HR@5$\uparrow$ &
NDCG@5$\uparrow$\\
\midrule
IntBMoE & 2.6802 & 14.5878 & 0.6852 & 0.8692 & 0.7850\\
\midrule
1-Layer & 2.7174 & 15.1410 & 0.6824 & 0.8680 & 0.7828\\
Fixed $\lambda$ & 2.6938 & 14.7871 & 0.6844 & 0.8685 & 0.7827\\
w/o Gate & 2.7660 & 15.8947 & 0.6820 & 0.8676 & 0.7828\\
w/o Shared & 2.7281 & 15.3035 & 0.6846 & 0.8689 & 0.7844\\
w/o Filter & 2.6829 & 14.6277 & 0.6847 & 0.8687 & 0.7846\\
Softmax Coeff. & 2.7071 & 14.9864 & 0.6835 & 0.8686 & 0.7839\\
\bottomrule
\end{tabular*}
\end{table*}

Table~\ref{tab:ablation-generalization}
reports component ablations on both MiniPile and IntTravel. The gate path has the largest impact on MiniPile and is among the most
influential components on IntTravel. The parameter-matched 1-Layer variant
also underperforms the full model, showing that the two-layer block structure
provides benefits beyond expert parameter count. The shared expert yields
smaller but consistent gains. Fixing $\lambda$ and removing feature filtering
hurt both autoregressive tasks, while the unconstrained, variance-scaled
coefficients consistently outperform softmax-normalized composition. These
results confirm that the main design choices remain effective beyond vision.

\paragraph{Online evaluation and deployment.}
We further evaluate cached IntBMoE in AMap's POI recommendation service for
the initial map screen displayed when a user opens the app. The production generative
recommender predicts the Top-10 POIs that the user is likely to visit and uses
them to determine the map viewport and zoom level. Because this service
requires a response within $60\,\mathrm{ms}$, we precompute and cache the
input-independent composed block parameters, removing block synthesis from the
request-time path.

We conducted a one-week online A/B experiment in which the control group used
the existing production model without an MoE module and the treatment group
used cached IntBMoE. The experiment handled approximately 5,000 queries per
second. Running on Alibaba T-Head PPUs, cached IntBMoE achieved an average
latency of $19\,\mathrm{ms}$ and a P99 latency of $38\,\mathrm{ms}$, remaining
within the serving budget while delivering a 2.4\% relative improvement in
UVCTR. Following the experiment, IntBMoE was fully deployed to serve
production traffic.

\section{Conclusion}

Existing MoE designs couple expert participation, execution, and
materialization: sparse routing limits participation, dense output mixing
increases execution, and parameter merging increases materialization. We
introduced IntBMoE to resolve this coupling,
allowing every token to benefit from the full expert pool without dense
execution or unbounded parameter materialization. IntBMoE separates reusable
block construction from token-level execution. Each candidate block composes
the full pool of expert bases, while Top-$k$ routing executes only a few blocks
per token, and a finite codebook bounds the number of materialized blocks
independently of the number of routing decisions. Within each block, DPRG
couples independently composed value and gate paths through residual
multiplicative interactions.
On the primary ImageNet-1K benchmark, IntBMoE consistently outperforms the
evaluated sparse and dense-participation MoE baselines. Analysis on ImageNet-1K
shows that different image classes induce distinct block-allocation
profiles. It also shows that blocks within each layer learn distinct
expert-composition recipes. Auxiliary experiments on MiniPile and IntTravel
further demonstrate generalization to language modeling and sequential
recommendation. For efficient serving, the composed blocks can be
cached to remove block synthesis from the request-time path. The full
deployment of IntBMoE in AMap's
latency-critical generative recommendation system further demonstrates its
practical utility.

%% file: tex/appendix.tex
\section{Detailed Experimental Configurations}
\label{app:experimental-configurations}

\subsection{Common Experimental Protocol}
All experiments use eight Alibaba T-Head PPUs, each with 96 GB of memory.
Within each dataset, all methods share the backbone, input processing, task
head, optimization schedule, seed set, and expert initialization; only the
FFN or MoE module differs. ImageNet-1K uses seeds 214797, 531770, and 635451,
MiniPile uses seeds 1, 145306, and 145849, and IntTravel uses seeds 42,
4981213, and 211231244.

Table~\ref{tab:training-config-appendix} summarizes the model dimensions,
input sizes, and training lengths. Here, $H$ and $d$ denote the number of
Transformer layers and the hidden size. $n_{\mathrm{heads}}$,
$d_{\mathrm{head}}$, and $d_r$ denote the number of attention heads, the head
dimension, and the router hidden size. $E$ is the number of expert bases in
each IntBMoE internal layer and the routed-expert count for baselines with a
shared path. BS, Input, and $\lambda_0$ denote the per-device batch size, the
image resolution or maximum sequence length, and IntBMoE's initial residual
scale, respectively.

\begin{table*}[t]
\centering
\small
\caption{Dataset-specific model and training configurations.}
\label{tab:training-config-appendix}
\renewcommand{\arraystretch}{1.05}
\setlength{\tabcolsep}{3pt}
\begin{tabular*}{\textwidth}{@{\extracolsep{\fill}}lccccccccl@{}}
\toprule
Dataset & $H$ & $d$ & $n_{\mathrm{heads}}\!\times\!d_{\mathrm{head}}$ &
$d_r$ & $E$ & BS & Input & $\lambda_0$ & Training\\
\midrule
ImageNet-1K & 8 & 192 & $3\!\times\!64$ & 64 & 16 & 256 &
$224^2$ ($p=16$) & 1 & 200 epochs\\
MiniPile & 18 & 768 & $12\!\times\!64$ & 64 & 8 & 8 & 1024 & 1 &
1 epoch\\
IntTravel & 8 & 96 & $1\!\times\!96$ & 64 & 32 & 64 & 126 & 0 &
610K steps\\
\bottomrule
\end{tabular*}
\end{table*}

\subsection{Dataset-Specific Configurations}

\paragraph{ImageNet-1K.}
We use a DeiT-Tiny-style
backbone~\cite{touvron2021training} shortened to eight Transformer layers. MoE
modules replace the FFNs in alternating layers (0, 2, 4, and 6) for patch
tokens, while the class token uses a separate dense FFN. We use AdamW with
$(\beta_1,\beta_2)=(0.9,0.999)$ and weight decay 0.05. The learning rate warms
up from $10^{-6}$ to $5\times10^{-4}$ over five epochs and then follows cosine
decay to $10^{-5}$. Standard dropout, attention dropout, patch dropout, and
position dropout are disabled, while stochastic depth uses a maximum DropPath
rate of 0.1. The augmentation pipeline applies a $224\times224$ bicubic random
resized crop with scale $[0.08,1.0]$ and aspect ratio $[0.75,1.33]$, random
horizontal flipping with probability 0.5, RandAugment using
two operations per image, magnitude 9, and a magnitude standard deviation of
0.5, and pixel-mode random erasing with probability 0.25 and one erased
region. Mixup
($\alpha=0.8$) and CutMix ($\alpha=1.0$) are applied in batch mode with a
combined probability of 1.0 and equal selection probabilities, and label
smoothing is 0.1.

\paragraph{MiniPile.}
We use a Llama-style causal
Transformer~\cite{touvron2023llama} and replace the FFN in every layer with an
MoE module. We train for one epoch on 1.523 billion tokens and evaluate on a
15.0-million-token test split. Text is tokenized
with a byte-level BPE vocabulary of 32,000. Four-step gradient accumulation gives an
effective batch size of 256 sequences across eight devices. We use AdamW with
$(\beta_1,\beta_2)=(0.9,0.95)$ and weight decay 0.1. The learning rate warms up
to $4\times10^{-4}$ over the first 10\% of training and then decays
cosinusoidally to $4\times10^{-5}$.

\paragraph{IntTravel.}
We use a causal Transformer for sequential recommendation
and replace every FFN and the prediction head with MoE modules. We use AdamW
with $(\beta_1,\beta_2)=(0.9,0.999)$, weight decay $10^{-6}$, and a constant
learning rate of $10^{-4}$.

\subsection{MoE Configurations and Fair Comparison}

\paragraph{Expert-parameter matching.}
We use the parameter count of the backbone's original Transformer FFN, which
is replaced by the MoE module, as the target for a single expert. Each internal
IntBMoE layer retains a separate pool of $E$ expert bases, and each basis
contains one weight matrix. For parameter accounting, we pair same-index bases
from the two internal layers and match their combined parameter count to this
target. For the baselines, one complete FFN or SwiGLU expert is matched to the
same target by counting all of its weight matrices. This matching sets the
input--intermediate--output dimensions of each two-layer IntBMoE block to
$d\!\rightarrow\!2844\!\rightarrow\!d$ on MiniPile and
$d\!\rightarrow\!4d\!\rightarrow\!d$ on ImageNet-1K and IntTravel. We also
match the always-active shared SwiGLU expert to the same
parameter target by adjusting its intermediate width. To keep the total expert
parameter count comparable across methods, a baseline with a shared path uses
$E$ routed experts and one shared expert, whereas a baseline without one uses
$E+1$ routed experts. For the CP and Tensor Ring variants of $\mu$MoE, we
choose the factorization ranks that make the total trainable parameter count
of the full model as close as possible to that of IntBMoE. Specifically, we
solve the CP rank $R$ under this parameter budget; for Tensor Ring, we fix
$R_1=R_2=4$ and solve $R_3$.

\paragraph{Baseline configurations.}
Switch Transformer follows its original Top-1 routing rule. Other Top-$k$
token-choice baselines use Top-2 routing when their original configurations do
not specify $k$. Expert Choice retains expert-to-token routing with a
capacity factor of 2. $\mu$MoE uses $\mathrm{entmax}_{1.5}$ routing, and MASS
uses Top-$p$ routing with $p=0.5$.

Soft MoE uses one slot per image token on ImageNet-1K. On MiniPile and
IntTravel, we make Soft MoE causal by constructing each slot only from the
current and preceding tokens. We use one slot per expert to limit computation.
SMEAR uses example-level
composition for images and token-level composition for autoregressive tasks.
Lory uses a segment size of 16.

For DynMoE and MASS, the initial expert-pool sizes on IntTravel, ImageNet-1K,
and MiniPile are 16, 8, and 4, respectively. Their maximum pool sizes match
the expert counts used by the other methods. We create all expert parameters
in advance up to these maxima. DynMoE initializes each expert's threshold to
correspond to a selection probability of 0.5 and uses a routing temperature of
1.0.
Every 300 optimizer steps, it removes unused experts and opens a new expert
when some tokens activate no expert. MASS detects semantic drift over the previous
200 optimizer steps and expands the pool using significance and similarity
thresholds of 0.01 and 0.001. Its redundancy regularizer has a coefficient of
0.01.

\paragraph{IntBMoE configuration.}
Each module contains $K=8$ codebook blocks and routes each token to $k=2$
blocks. The block-embedding dimension, router hidden dimension, and
hypernetwork hidden dimension are 32, 64, and 16, respectively. Each block
contains two internal DPRG layers. These layers use equally sized,
layer-specific expert pools and share
the same pair of value and gate composition coefficients. An always-active
SwiGLU expert provides the shared path. We initialize the expert matrices with
PyTorch's default \texttt{nn.Linear} scheme and set all expert biases to zero.
This small initial weight scale limits perturbations from the composed
transformations early in training.